\documentclass{article}
\usepackage{spconf,amsmath,epsfig}
\usepackage{amsfonts}
\usepackage{hyperref}
\usepackage{ulem}
\usepackage{booktabs}
\let\OLDthebibliography\thebibliography
\renewcommand\thebibliography[1]{
  \OLDthebibliography{#1}
  \setlength{\parskip}{0pt}
  \setlength{\itemsep}{0pt plus 0.3ex}
}
\usepackage{color}
\usepackage[binary-units = true]{siunitx} % use this package module for SI units
\usepackage{url}
\usepackage{multirow}

\begin{document}\sloppy

% Example definitions.
% --------------------
\def\x{{\mathbf x}}
\def\L{{\cal L}}

% Title.
% ------
\title{SAM-Lightening: A Lightweight Segment Anything Model with Dilated Flash Attention to Achieve 30$\times$ Acceleration}
%
% \name{Yanfei Song$^{1,2}$, Bangzheng Pu$^{1,2}$, Peng Wang$^{1,2}$, Dong Dong$^{1,2}$, Hongxu Jiang$^{1,2}$\sthanks{Corresponding Author: jianghx@buaa.edu.cn}, Yiqing Shen$^{3}$}
% \address{
% $^{1}$Beihang University \\
% $^{2}$Beihang Hangzhou Innovation Institute \\
% $^{3}$Johns Hopkins University
% }
\maketitle

\begin{abstract}
Segment Anything Model (SAM) has garnered significant attention in segmentation tasks due to their zero-shot generalization ability.
However, a broader application of SAMs to real-world practice has been restricted by their low inference speed and high computational memory demands, which mainly stem from the attention mechanism.
Existing work concentrated on optimizing the encoder, yet has not adequately addressed the inefficiency of the attention mechanism itself, even when distilled to a smaller model, which thus leaves space for further improvement.
In response, we introduce SAM-Lightening, a variant of SAM, that features a re-engineered attention mechanism, termed Dilated Flash Attention. 
It not only facilitates higher parallelism, enhancing processing efficiency but also retains compatibility with the existing FlashAttention.
Correspondingly, we propose a progressive distillation to enable an efficient knowledge transfer from the vanilla SAM without costly training from scratch.
Experiments on COCO and LVIS reveal that SAM-Lightening significantly outperforms the state-of-the-art methods in both run-time efficiency and segmentation accuracy.
Specifically, it can achieve an inference speed of 7 milliseconds (ms) per image, for images of size 1024$\times$1024 pixels, which is $30.1\times$ faster than the vanilla SAM and $2.1\times$ than the state-of-the-art. 
Moreover, it takes only \SI{244}{\mega\byte} memory, which is $(3.5\%)$ of the vanilla SAM.
The code and weights are available at \url{https://anonymous.4open.science/r/SAM-LIGHTENING-BC25/}.
% 
% \textcolor{blue}{check the link to the code here}
\end{abstract}
\begin{keywords}
Segment Anything Model (SAM), 
Knowledge Distillation,
Computational Efficient Attention Mechanisms
\end{keywords}

\section{Introduction}
\label{sec:intro}

\begin{figure*}[ht]
    \centering
    \includegraphics[width=\linewidth]{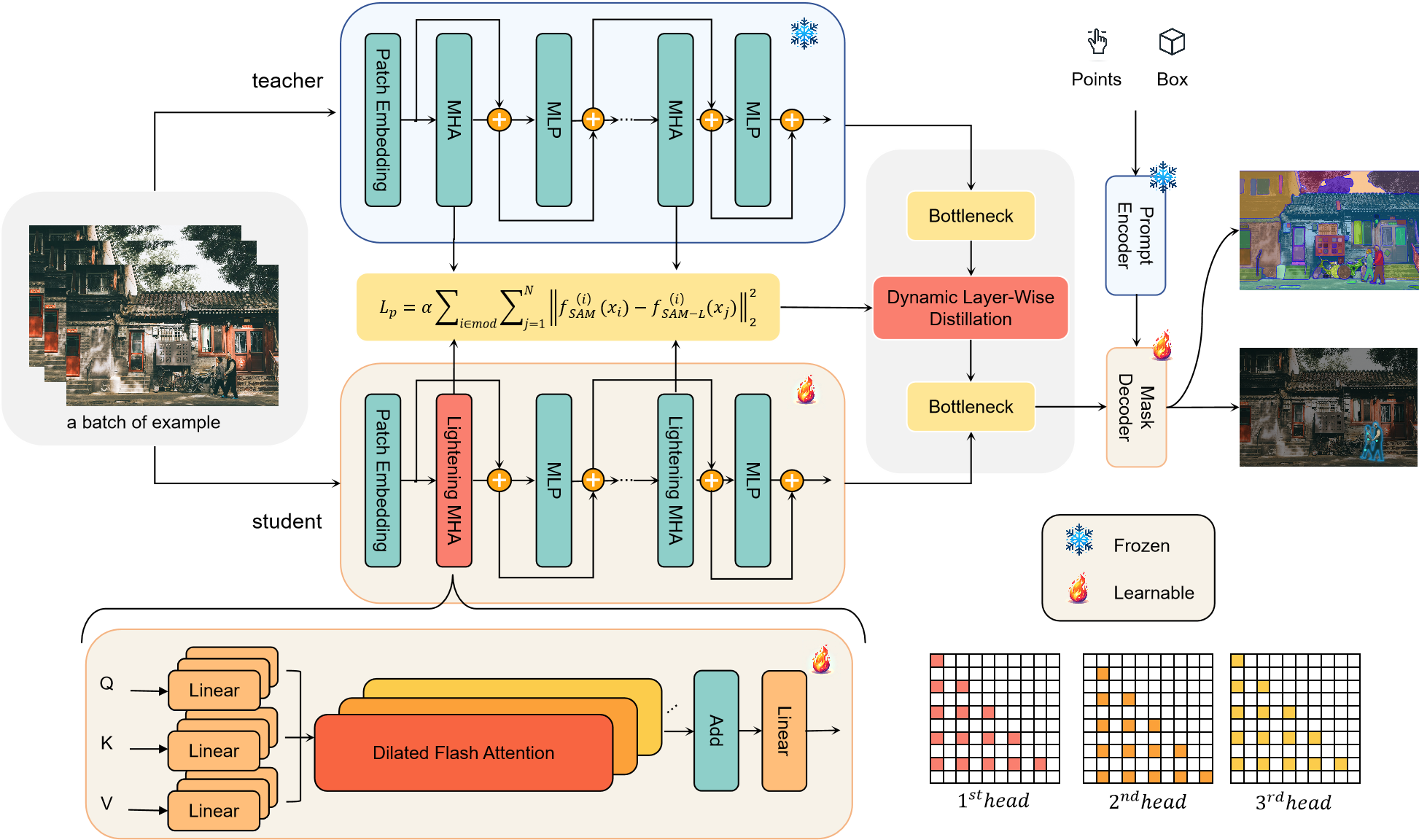}
    \caption{The overall framework of SAM-Lightening along with the dynamic layer-wise distillation that can efficiently transfer knowledge from the vanilla SAM without training from scratch.}
    \label{fig:Lightening Architecture with DLD}
\end{figure*}
Image segmentation has been traditionally constrained by the necessity for deep learning models to be specifically trained on datasets designed for particular tasks.
This specialization of hand-crafted datasets often limits their generation ability.
Addressing this constraint, the Segment Anything Model (SAM)~\cite{kirillov2023segment} represents a paradigmatic shift with its zero-shot learning abilities that allow itself to segment new and unseen images.
However, SAM's application in varied sectors like augmented reality (AR), image editing, deployment on smartphones and medical imaging~\cite{Archit_Nair_Khalid_Hilt_Rajashekar_Freitag_Gupta_Dengel_Ahmed_Pape_2023, Ma_Wang_2023, cheng2023sam, Yang_Gao_Li_Gao_Wang_Zheng_2023, shen2023anything} is impeded by its computational burden challenge in its image encoder, which comprises a substantial 632 million parameters.
This size is roughly 20 times that of conventional segmentation networks like U-Net~\cite{Ronneberger_Fischer_Brox_2015}, leading to high computational demands.
% 
% This aspect becomes a critical hindrance in resource-constrained devices, such as AR glasses and smartphones, underscoring a need for optimization.

In response to this challenge, various efforts have been initiated.
For example, FastSAM~\cite{zhao2023fast} adopts a strategy of replacing SAM's transformer encoder with a more streamlined convolutional neural network (CNN), aiming to create a lighter model.  
However, this often leads to diminished accuracy, especially in complex segmentation tasks. 
Another notable approach is MobileSAM~\cite{zhang2023faster}, which employs distillation techniques to transfer knowledge from SAM's encoder to a more compact ViT-tiny~\cite{2022TinyViT} encoder.
Similarly, initiatives like EfficientSAM~\cite{xiong2023efficientsam} aim to refine the training processes of MobileSAM to improve accuracy.
Conversely, SAMFast~\cite{PyTorch2023Accelerating} focuses on speed optimization of the original SAM through techniques such as quantization and pruning, but these modifications have limited impact on performance enhancement.

Our research identifies key limitations in previous works~\cite{zhang2023faster, xiong2023efficientsam, PyTorch2023Accelerating} on SAM, primarily in terms of inefficient computation and memory usage in attention mechanisms. To address these issues, we integrate FlashAttention~\cite{dao2022flashattention} and dilated attention mechanisms into our SAM framework, providing orthogonal improvements over existing methods. These enhancements not only reduce memory consumption but also improve parallel processing, making them complementary to previous advancements.
% Our research diverges from these methods by delving into a more nuanced optimization of the encoder, particularly targeting the attention operator within the model.
% % 
% We introduce an advanced SAM structure that incorporates the FlashAttention~\cite{dao2022flashattention} mechanism, which can reduce memory usage and boost parallel processing efficiency. 
% % 
% FlashAttention's effectiveness hinges on the continuous, dense computation of the attention matrix, a feature not fully exploited in previous works for SAMs~\cite{zhang2023faster, xiong2023efficientsam, PyTorch2023Accelerating}.Due to our modifications to the attention mechanism and the reduction in model structure, decoupled distillation~\cite{zhang2023faster} methods have shown limited effectiveness.
However, directly applying these mechanisms to SAM would necessitate a complete retraining of the model, incurring substantial computational costs.  To circumvent this challenge, 
we proposed a dynamic layer-wise distillation (DLD). DLD implements a progressive distillation scheme for the image encoder by progressively allocating feature weights, effectively facilitating the transfer of knowledge from SAM to our lightweight model.
We demonstrate that our model (SAM-Lightening) is not only expressive enough to represent the original SAM but is also computationally efficient, completing inference within \SI{7}{\milli\second}.

In brief, our main contributions are four-fold:
\begin{itemize}
    \item We introduce a novel SAM structure, SAM-Lightening, to significantly reduce the computational complexity. 
    \item We design a novel dilated flash attention mechanism to replace the vanilla self-attention to enhance the efficiency and inference speed of SAM-Lightening.
    \item To efficiently transfer the knowledge from vanilla SAM to SAM-Lightening, we propose a dynamic layer-wise distillation without compromising the performance.
    \item SAM-Lightening achieves state-of-the-art performance of 7 ms per image, which is $30.1\times$ faster than vanilla SAM.
\end{itemize}
% To facilitate the usage of SAM-Lightening for further research, we provide codes and all the associated weights. 

\section{Related work}
\noindent\textbf{Segment Anything Model:} 
SAM comprises three main parts: the image encoder, prompt encoder, and mask decoder. 
Notably, the image encoder is the most parameter-intensive segment of SAM, accounting for a substantial 98.3\% of its processing time~\cite{kirillov2023segment}, which highlights the need for optimization.
FastSAM~\cite{zhao2023fast} employs a CNN encoder, specifically the YOLOv8-seg~\cite{yolov8_ultralytics}, to replace the ViT encoder to enhance processing speed. 
However, it has been observed to compromise segmentation precision, particularly in complex scenarios and in capturing fine edge details.
MobileSAM~\cite{zhang2023faster} distill the encoder to reduce both the model size and computational requirements. 
Nevertheless, the imbalance in MobileSAM's encoder structure and parameter distribution limits its potential for practical deployment and performance optimization. 
SAMFast~\cite{PyTorch2023Accelerating} represents another optimization strategy, focusing on enhancing the processing speed of SAM using methods like quantization and sparsification. While this scheme does offer some acceleration, its overall impact remains moderate. 
EfficientSAM~\cite{xiong2023efficientsam}, on the other hand, improves upon MobileSAM's training methodology, specifically targeting the accuracy aspect of the MobileSAM approach.

\noindent\textbf{FlashAttention:} 
The FlashAttention mechanism~\cite{dao2022flashattention} introduces an efficient and accurate approach for computing attention in neural networks. 
It achieves a significant reduction in high bandwidth memory reads and writes, primarily through strategic tiling and recomputation techniques.
Building upon this, FlashAttention-2~\cite{dao2023flashattention} further refines the process by enhanced matrix multiplication operations. 
These improvements have been shown to deliver up to a twofold increase in performance in specific computational settings.

\noindent\textbf{Knowledge Distillation:}
Knowledge distillation~\cite{Hinton_Vinyals_Dean_2015} is a technique for transferring knowledge from a complex model to a simpler one. 
They aim to retain the performance attributes of the larger model while significantly reducing its computational footprint and model size.
MobileSAM employs a decoupled knowledge distillation by extracting outputs from the original SAM's ViT-H image encoder and using them to distill into a pre-trained ViT-tiny encoder directly. 
This strategy proves particularly beneficial for smaller models that already possess pre-trained parameters.

\section{Methods}
\subsection{Dilated Flash Attention}
To address the high computational demands in the image encoder of SAM, we design a novel attention operation with FlashAttention to expedite the inference speed.

\noindent\textbf{Segmentation and Sparsification}: 
To alleviate the computational burden in processing ($Q, K, V$) in attention operation, we divide each input into equal-length parts ($w$) and then apply sparsification along the sequence dimension within each segment.
This sparsification involves selecting rows at fixed intervals ($r$), thereby reducing the volume of data the attention mechanism needs to process.
As shown in Fig.~\ref{fig:Lightening Architecture with DLD}, the sparsification process can be formulated as:
\begin{equation}
    \widetilde{X}_i = [X_{iw}, X_{iw+r}, X_{iw+2r}, \ldots, X_{(i+1)w-1}],
\end{equation}
 Here, $\widetilde{X}_i$ represents the sampled sparse matrix. $X$ represents any of the variables $Q$, $K$, or $V$.

\noindent\textbf{Parallel Processing With FlashAttention}: 
Sparsified segments of each input data are dense matrices that can participate in the attention calculation independently and thus can be processed in parallel. 
This parallelism is vital for efficiently managing large-scale image datasets, significantly speeding up the processing time and enhancing the efficiency of our model for real-time image segmentation.
Incorporating FlashAttention further increases efficiency by parallelizing dense matrix computations in the process.

\noindent\textbf{Output Recomposition}: 
In the proposed Dilated Flash Attention framework, we process sparsified segments in parallel, implementing a softmax function applied to the product of $\widetilde{Q}_i$ and the transpose of $\widetilde{K}_i$, subsequently followed by multiplication with $\widetilde{V}_i$ as follows:
\begin{equation}
\widetilde{O}_i = \texttt{softmax}(\widetilde{Q}_i \cdot \widetilde{K}^T_i ) \cdot \widetilde{V}_i.
\end{equation}
The reassembly of these outputs into the cohesive final output $O$ involves a meticulously designed process:
\begin{enumerate}
    \item[(1)] Initially, we establish a zero matrix $O_{\text{init}}$ that mirrors the dimensions of the original input for accumulating the outputs of the individual segments.
    \item[(2)] For each computed segment output $\widetilde{O}_i$, a specific offset $\gamma_i$ is identified. This offset determines the precise starting position of $\widetilde{O}_i$ within the $O_{\text{init}}$ matrix.
    \item[(3)] Each \( \widetilde{O}_i \) is mapped to \( O_{\text{init}} \) using a mapping operation based on its $\gamma_i$:
    \begin{equation}
    O = \sum_{i} \texttt{MAP}(O_{\text{init}}, \widetilde{O}_i, \gamma_i)
    \end{equation}
\end{enumerate}
The ``\texttt{MAP}'' operation places each \( \widetilde{O}_i \) element into \( O_{\text{init}} \) according to the position determined by \( \gamma_i \). This guarantees the accurate alignment of each segment's output within the final output matrix \( O \), based on its original input position.

\noindent\textbf{Computation Efficiency}
With the proposed Dilated Flash Attention mechanism, efficiency is quantitatively enhanced by a factor of $\frac{N}{wr^2}$, where $N$ represents the total size of the input, $w$ the length of each segment, and $r$ the interval of sparsification.
This mathematical relationship demonstrates that Dilated Flash Attention requires substantially fewer computations for any given input size. 
Consequently, this boosts the model’s capability in efficiently processing large-scale image segmentation tasks, marking a notable improvement in both performance and practicality.

\subsection{Dynamic Layer-Wise Distillation (DLD)}
Training the SAM-Lightening from scratch is costly, while layer adaptation is challenging due to the distinctive structures between SAM with ViT-H as the feature encoder and SAM-Lightening. 
To enable efficient knowledge transfer from vanilla SAM to the proposed framework, we propose a novel Dynamic Layer-Wise Distillation (DLD), which dynamically modifies feature weights to enhance the layer-wise distillation between the models~\cite{Ji_Heo_Park_2022}.

\noindent\textbf{Dynamic Layer-Wise Weights: }
When preceding layers are not well-distilled, the performance of subsequent layers can suffer from low-quality features extracted from preceding layers. By assigning greater weight to the loss of these initial layers, dynamic weighting ensures they receive more focus during the training process. This helps in better aligning the student model with the teacher model in the initial stages. 
Given a deep neural network consisting of $L$ layers, each layer $i$ is associated with a temporal weight $\alpha_i(t)$. This mechanism adjusts the significance of each layer $i$ in the neural network across various training stages $t$.
The initial layer retains maximum emphasis ($ \alpha_1(t)=1$) and the subsequent layers adhere to a dynamic weighting scheme, which can be mathematically represented by the piece-wise function:
\begin{eqnarray}
    \alpha_i(t) = 
     \begin{cases} 
      0 & \text{for } t < T_i \\
      \frac{t - T_i}{\Delta t} & \text{for } T_i \leq t < T_i + \Delta t \\
      1 & \text{for } t \geq T_i + \Delta t 
     \end{cases}
\end{eqnarray}
Where $T_i$ denotes the epoch at which the $i^{th}$ layer commences updating its weight, and the previous layer has reached saturation, i.e., $T_i=T_{i-1}+\Delta t$. The parameter $\Delta t$ captures the number of epochs over which the weight transitions from 0 to 1.
For a predefined epoch increment $\Delta t$, each layer sequentially activates its learning potential after the preceding layer reaches its peak weight. This mechanism facilitates a cascading knowledge absorption from the teacher model.

\noindent\textbf{Decoupled Feature Distillation}: The distillation process transfers knowledge from SAM's encoder (the teacher model) to our proposed encoder (the student model), as shown in Fig.\ref{fig:Lightening Architecture with DLD}. We have chosen the $N$ layers closest to the output for feature distillation. Since these deeper layers are directly related to the model's outputs, distilling them can more effectively transfer crucial information for prediction results. These layers are designated as ``Focus Layers''.

During the initial phase of training, layers closer to the input are given precedence. Here, the intent is to align the SAM-Lightning primary feature representations of the student model, expressed as $f_{\text{SAM-L}}^{i}(x)$, with those of the teacher model, $f_{\text{SAM}}^{i}(x)$, for the $i$ layers closest to the input.
As training advances, the layer-wise weighting dynamically shifts. The loss associated with subsequent layers is incrementally amplified. In the progress, the loss function evolves to assimilate representations from succeeding layers:
\begin{eqnarray}
    L_{\text{P}} =  \sum_{i \in \text{Focus}} \alpha_i(t) \sum_{j=1}^{N} \left\| f_{\text{SAM}}^{(i)}(x_j) - f_{\text{SAM-L}}^{(i)}(x_j) \right\|_2^2
\end{eqnarray}
where $L$ is the complete count of layers, and the coefficient $\alpha(i)$ is a piece-wise function determined by the training epoch and the layer $i$. The integrated distillation loss is formulated as:
\begin{eqnarray}
    L_{\text{integrated}} = L_{\text{P}} + \lambda L_{\text{output}}
\end{eqnarray}
where \( L_{\text{P}} \) encapsulates the weighted sum of all selected feature layer losses, \( L_{\text{output}} \) is the loss for the image encoder output layer, and \( \lambda \) is a scaling factor to balance the significance of the decoder output in the overall distillation process.

\noindent\textbf{Align Decoder}: Additionally, the lightweight image encoder obtained through decoupled distillation has alignment issues with the frozen decoder, especially for point-based prompt segmentation tasks. Therefore, we fine-tuned the decoder by sampling point prompts and box prompts on the SA-1B dataset to align with the image encoder.
The loss function is defined as follows:
\begin{eqnarray}
    L_{\text{fine-tune}} = 20 \times \text{IOU} + \text{Dice} + \text{Focal Loss}
\end{eqnarray}
Here, IOU represents the Intersection over Union loss, while Dice loss and Focal Loss are used to address class imbalance and challenging segmentation regions, respectively.

\section{Experiment}
\subsection{Experimental Setups}
Our model utilizes $1\%$ of the SA-1B dataset for distillation and fine-tuning. 
It features an encoder with an embedding dimension of 384, six attention heads, and a six-layer structure. 
For the FlashAttention component, we use bfloat16. 
Both the distillation and fine-tuning processes are conducted for 10 epochs each, with a learning rate of $10^{-3}$ and a batch size of 32. 
Gradient accumulation is set with a step size of 4. 
The model is trained on two NVIDIA RTX 4090 GPUs. 
To enhance training speed, the outputs of SAM's image encoder are saved~\cite{2022TinyViT,zhang2023faster}.

\subsection{Results}
%\subsubsection{Run-Time And Memory Efficiency Evaluation} 
\noindent\textbf{Run-Time And Memory Efficiency Evaluation:}
We compare the performance of our proposed SAM-Lightening with vanilla SAM (\textit{i.e.,} SAM-ViT-H)~\cite{kirillov2023segment}, FastSAM~\cite{zhao2023fast}, MobileSAM~\cite{zhang2023faster}, EfficientSAM~\cite{xiong2023efficientsam}, SAMFast~\cite{PyTorch2023Accelerating} in Table~\ref{table:1} and Table~\ref{table:2}.
Regarding the segmentation performance, the vanilla SAM is considered as the upper bound.
Importantly, Table~\ref{table:1} shows that SAM-Lightening outperforms all its counterparts in terms of inference latency and peak memory usage, achieving $30.1\times$ acceleration, $96.5\%$ peak memory reduction when compared to vanilla SAM, and $2.1\times$ acceleration when compared to the state-of-the-art.
The throughput comparison in Table~\ref{table:2} further reinforces SAM-Lightening's superior performance, which achieves the highest throughput across various batch sizes. 
Conclusively, this high throughput with its low latency and memory usage, positions SAM-Lightening as a highly efficient model for image segmentation tasks.

\begin{table}[t!]
\centering
\caption{Performance comparison on Nvidia RTX 4090 GPU, where ``Enc.'' refers to the Encoder, ``Dec.'' to the Decoder, ``Mem.'' to Memory usage, ``Tot.'' to Total Time, and ``SU'' denotes the Speed-Up ratio.}
\begin{tabular}{p{2.6cm}p{0.6cm}p{0.6cm}p{0.6cm}p{0.8cm}p{1cm}}
\toprule
\textbf{Model} & \textbf{Enc.} ms & \textbf{Dec.} ms & \textbf{Tot.} ms & \textbf{S.U.} & \textbf{Mem.}\\
\midrule
SAM-ViT-H          & 216.1       & 3.8        & 219.9       & 1.0$\times$           & 5.7GB \\
\midrule
SAMFast        & 23.2         & 3.8       & 27.0        & 8.5$\times$         & 4.1GB \\
FastSAM      & 20.7        & 3.4           & 24.1       & 9.1$\times$         & 2.6GB \\
EfficientSAM   & 22.3         & 3.8       & 26.1        & 8.3$\times$         & 309MB \\
MobileSAM      & 8.1         & 3.8       & 11.9        & 18.5$\times$         & 309MB \\
\midrule
\textbf{SAM-Lightening}  & \textbf{3.5}         & \textbf{3.4}       & \textbf{6.9}        & \textbf{30.1$\times$}         & \textbf{224MB} \\
\bottomrule
\end{tabular}
\label{table:1}
\end{table}

\begin{table}[b!]
\centering
\caption{Parallel throughput comparison. Inference times are given in milliseconds (ms).}
\begin{tabular}{p{2.6cm}p{0.9cm}p{0.9cm}p{0.9cm}p{1.2cm}p{1.5cm}}
\toprule
\textbf{Model} & \textbf{Size 1}& \textbf{Size 4}  & \textbf{Size 8}  & \textbf{Size 16}  \\
\midrule
SAM-ViT-H          & 219.9       & 944.9       & OOM       & OOM       \\
\midrule
SAMFast        & 53.6       & 206.6       & 438.2       & 964.2       \\
FastSAM      & 24.1       & 80.1       & 171.5       & 349.1       \\
EfficientSAM   & 22.3       & 79.2       & 157.7       & 317.5       \\
MobileSAM      & 8.1       & 34.1       & 72.3       & 156.8       \\
\midrule
\textbf{SAM-Lightening}  & \textbf{3.5}       & \textbf{13.0}       & \textbf{27.2}       & \textbf{59.2}       \\
\bottomrule
\end{tabular}
\label{table:2}
\end{table}

%\subsubsection{Point \& Box Performance Comparison} 
\begin{figure}[t!]
    \centering
    \includegraphics[width=0.98\linewidth]{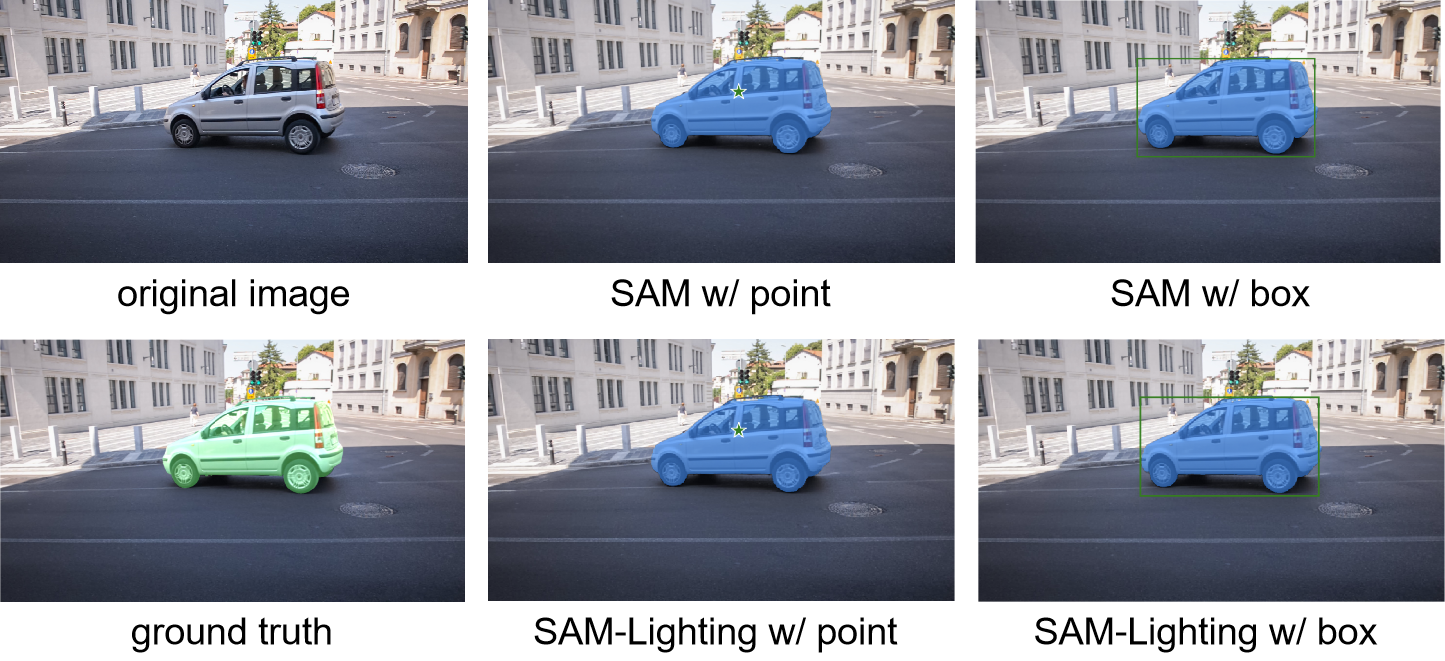}
    \caption{Representative image segmentation results between SAM-Lightening and the vanilla SAM in prompt mode.}
    \label{fig:my_label}
\end{figure}

\noindent\textbf{Comparison In Box/Point Prompt Mode:} 
We first evaluated the performance under bounding boxes and point-based prompts.
% 
% For bounding box prompts, we leveraged the ground-truth annotation in the COCO~\cite{lin2014microsoft} and LVIS~\cite{Gupta_Dollar_Girshick_2019} to synthesize prompts that define areas of interest in each image across all the methods.
For bounding box prompts, we followed the settings in vanilla SAM by leveraging the ground-truth annotation in the COCO~\cite{lin2014microsoft} and LVIS~\cite{Gupta_Dollar_Girshick_2019} to synthesize bounding boxes that define areas of interest in each image. 
% 
% This approach assesses the models' capability to segment objects within box prompts accurately.
% 
For point prompts, we randomly sampled points within the ground-truth masks from images, challenging all the models to accurately segment the object or region associated with each point. 
% 
% This method evaluates the precision of the models in segmenting based on point prompts.
% 
Quantitatively, we used mean Intersection over Union (mIoU) as the metric.
As shown in Table~\ref{table:segmentation_comparison}, both SAMFast and MobileSAM suffer from a performance decline when compared to vanilla SAM, particularly with point prompts. 
FastSAM, as a CNN-based model, shows an even more pronounced drop, which is especially evident in the handling LVIS dataset that contains a large number of small objects.
This observation reflects the limitations of CNN-based encoders in processing more complex segmentation scenarios. 
In contrast, SAM-Lightening matches the original SAM in terms of segmentation performance to the best context. 
This holds even in scenarios of point-based prompts, where SAM-Lightening achieves mIoU similar to the vanilla SAM.

\begin{table}[b!]
\centering
\caption{Segmentation performance comparison in terms of mIOU on COCO and LVIS. The labels ``Box'', ``1P'', and ``3P'' correspond to the use of a bounding box, one point, and three points as prompts, respectively.}
\resizebox{\columnwidth}{!}{%
\begin{tabular}{l|ccc|ccc}
\toprule
\multirow{2}{*}{\textbf{Model}} & \multicolumn{3}{c|}{\textbf{COCO}} & \multicolumn{3}{c}{\textbf{LVIS}} \\
\cline{2-7}
               & \textbf{Box} & \textbf{1P} & \textbf{3P} & \textbf{Box} & \textbf{1P} & \textbf{3P} \\
\midrule
SAM-ViT-H          & 80.1 & 49.2 & 72.5 & 83.8 & 60.6 & 74.7 \\
\midrule
SAMFast        & 77.3 & 44.7 & 66.3 & 80.5 & 54.9 & 69.4 \\
FastSAM      & 65.0 & 50.9 & 52.4 & 61.5 & 41.5 & 41.8 \\
EfficientSAM   & 77.8 & 43.6 & 69.7 & 79.5 & 53.7 & 72.9 \\
MobileSAM      & 77.9 & 47.9 & 67.4 & 78.5 & 55.4 & 66.8 \\
\midrule
\textbf{SAM-Lightening} & \textbf{78.8} & \textbf{48.4} & \textbf{72.5} & \textbf{81.0} & \textbf{59.9} & \textbf{74.6} \\
\bottomrule
\end{tabular}
}
\label{table:segmentation_comparison}
\end{table}

%\subsubsection{Anything Module Performance Comparison} 

\noindent\textbf{Comparison In Anything Mode:} 
\begin{figure}[h]
    \centering
    \includegraphics[width=\linewidth]{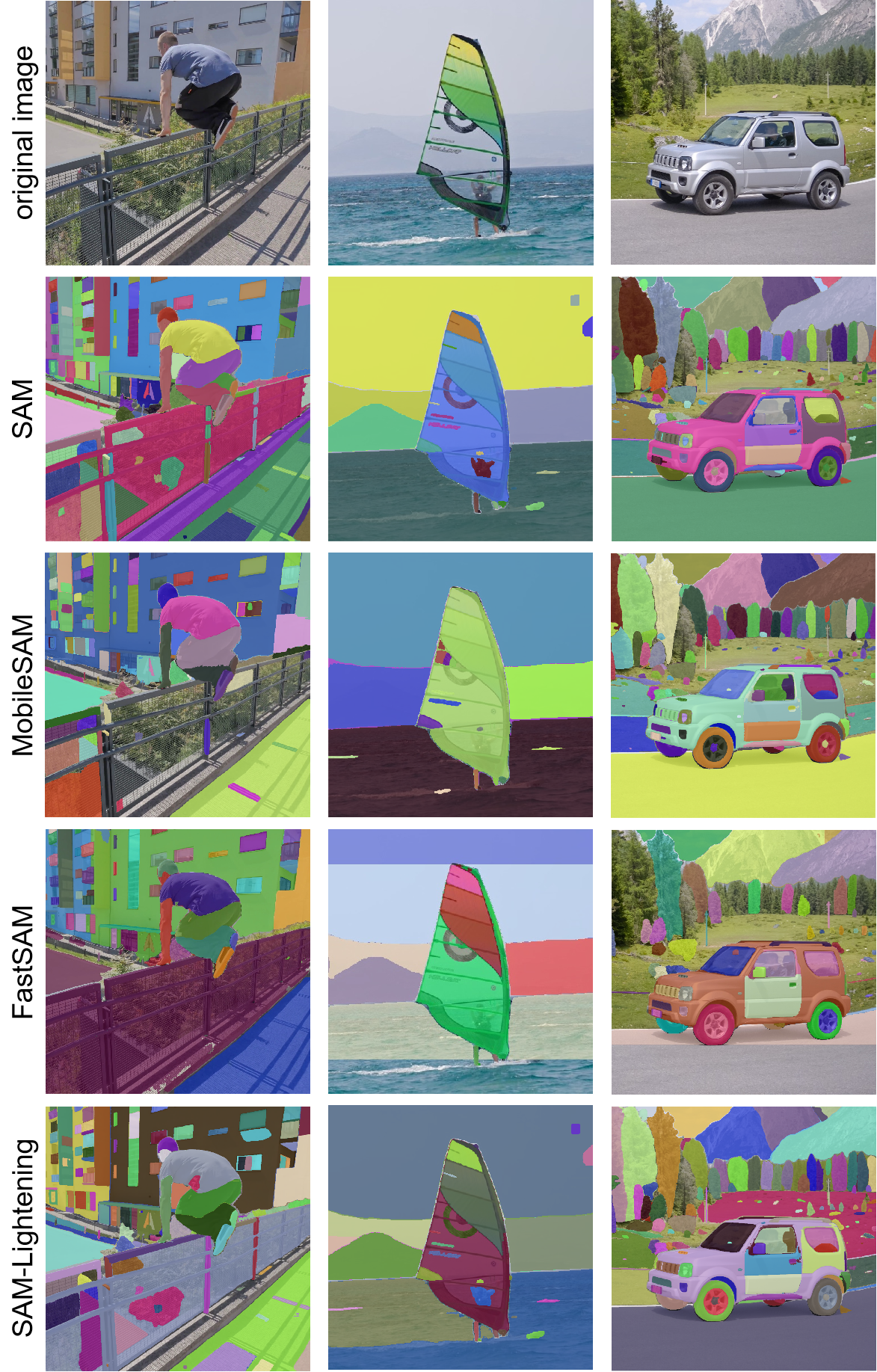}
    \caption{Representative samples under anything mode.}
    \label{fig: Segment Anything module Comparison}
\end{figure}
While the segment-anything mode is an innovative approach, it is not a commonly used segmentation method and thus does not effectively represent typical segmentation tasks. 
Therefore, our analysis has primarily focused on visually comparing the segmentation outcomes through point-based and box-based methods, which are more prevalent in practical applications. 
However, for completeness and to demonstrate the versatility of the models, we have also included the outputs of the segment-anything mode in our comparison.

From the representative samples demonstrated in Fig.~\ref{fig: Segment Anything module Comparison}, both SAM-Lightening and MobileSAM exhibit segmentation results that are nearly indistinguishable from those of the vanilla SAM. 
This similarity is notable in terms of edge clarity and detail preservation, which are hallmarks of high-quality segmentation. 
SAM-Lightening demonstrates its robustness and accuracy, aligning closely with the performance of the vanilla SAM.

\subsection{Ablation study}
\begin{figure}[t!]
    \centering
    \includegraphics[width=0.96\linewidth]{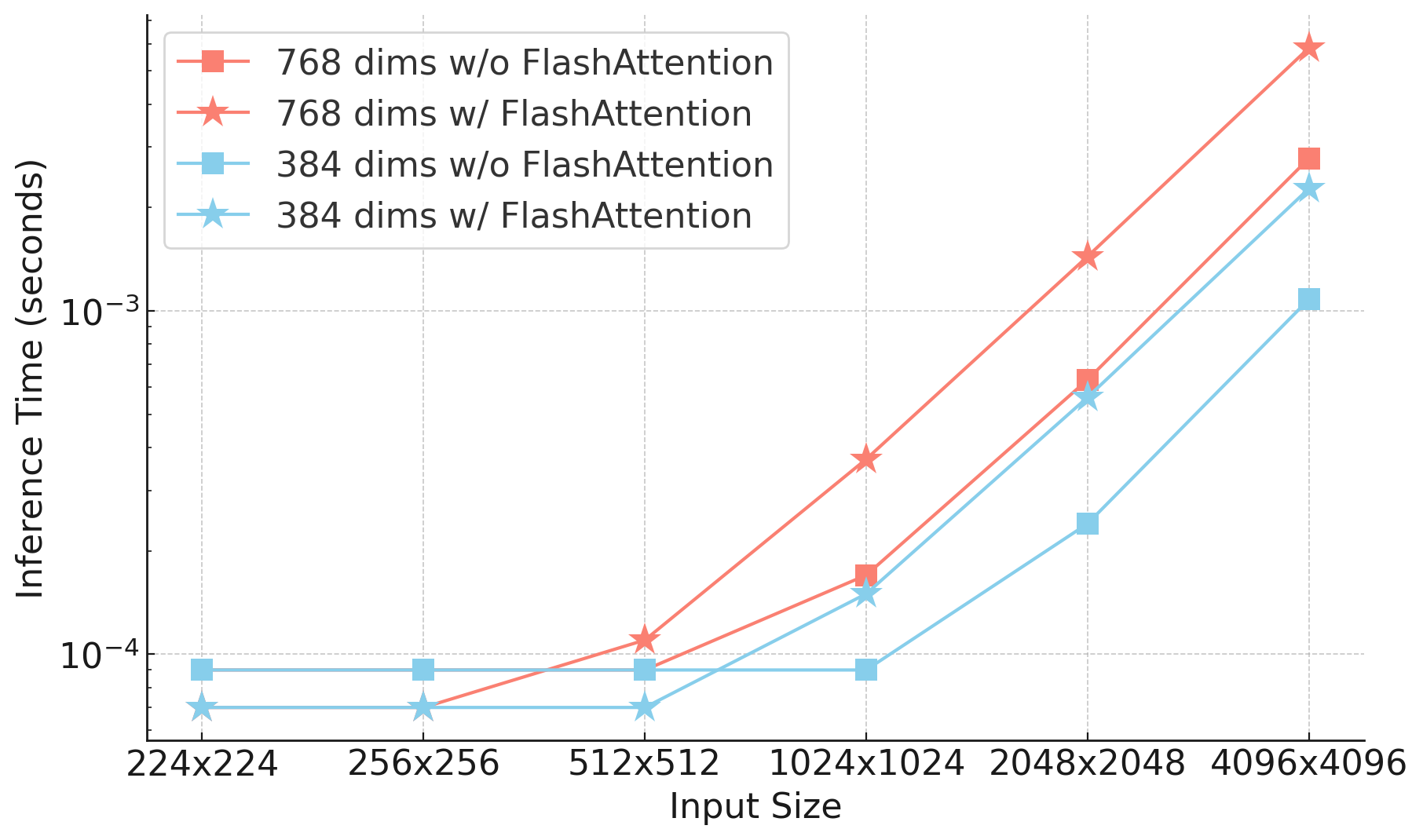}
    \caption{Impacts of inference time with FlashAttention over input size, where we select two embedding dimensions, namely 768 and 384, for comparison.}
    \label{fig:Flash}
\end{figure}
It's noteworthy that many previous works~\cite {cheng2023sam,mmdetection,wang2023seggpt} use smaller input sizes for SAM other than 1024. 
For a fair comparison, we also conducted experiments in these scenarios and found that keeping FlashAttention for input sizes equal to or smaller than $512\times512$ achieves optimal performance. 
This indicates that the applicability of FlashAttention depends on the model's input size and specific hardware configuration. 
The decision to use FlashAttention should be made based on the specific application context and performance requirements.
Although FlashAttention accelerates training in model distillation, its impact on inference performance is determined by various hardware metrics. On our inference platform, especially for the SAM with a 1024 input size, the multi-head attention operator exhibits a more computation-intensive characteristic. As shown in Fig.~\ref{fig:Flash}, this results in a slightly lower inference speed with FlashAttention compared to without it. Therefore, we opt to use FlashAttention during the distillation process to optimize performance while removing it during the evaluation phase.

\section{Conclusion}
We propose SAM-Lightening to address the primary limitations of high computational demand and slow inference speed in vanilla SAM to make it more suitable for deployment on resource-constrained devices.
Our approach involves the redesign of the image encoder in SAM, by distilling the self-attention operators into dilated flash attentions with dynamic layer-wise distillation. 
These optimizations contribute to a notable reduction in computational complexity and memory usage without compromising the segmentation performance.
Specifically, SAM-Lightening can complete inference within 7 milliseconds per image, achieving a $30.1\times$ speed up over SAM-ViT-H. 
Since SAM-Lightening is complementary to pruning and quantization, one future direction can look into the integration with them.

\bibliographystyle{IEEEbib}
\bibliography{icme2023template}

\end{document}